\documentclass[preprint,12pt]{elsarticle}

\usepackage{amsfonts}
\usepackage{amsmath}
\usepackage{bm}
\usepackage{graphicx}
\usepackage[dvipsnames]{xcolor}
\usepackage{booktabs}
\usepackage{subfigure}
\usepackage{xfrac}
\usepackage{threeparttable}
\usepackage{multirow}
\usepackage{rotating}
\usepackage{changepage}

\usepackage{lineno,hyperref}
\modulolinenumbers[5]

\journal{ISPRS J. Photo. Remote Sens.}

\begin{document}

\begin{frontmatter}

\title{RiFCN: Recurrent Network in Fully Convolutional Network for Semantic Segmentation of High Resolution Remote Sensing Images}


\author[firstaddress,secondaryaddress]{Lichao Mou}
\ead{lichao.mou@dlr.de}

\author[firstaddress,secondaryaddress]{Xiao Xiang Zhu\corref{correspondingauthor}}
\cortext[correspondingauthor]{Corresponding author}
\ead{xiao.zhu@dlr.de}

\address[firstaddress]{Remote Sensing Technology Institute (IMF), German Aerospace Center (DLR),  Oberpfaffenhofen, 82234 Wessling, Germany}
\address[secondaryaddress]{Signal Processing in Earth Observation (SiPEO), Technical University of Munich (TUM),  Arcisstr. 21, 80333 Munich, Germany}

\begin{abstract}
Semantic segmentation in high resolution remote sensing images is a fundamental and challenging task. Convolutional neural networks (CNNs), such as fully convolutional network (FCN) and SegNet, have shown outstanding performance in many segmentation tasks. One key pillar of these successes is mining useful information from features in convolutional layers for producing high resolution segmentation maps. For example, FCN nonlinearly combines high-level features extracted from last convolutional layers; whereas SegNet utilizes a deconvolutional network which takes as input only coarse, high-level feature maps of the last convolutional layer. However, how to better fuse multi-level convolutional feature maps for semantic segmentation of remote sensing images is underexplored. In this work, we propose a novel bidirectional network called recurrent network in fully convolutional network (RiFCN), which is end-to-end trainable. It has a forward stream and a backward stream. The former is a classification CNN architecture for feature extraction, which takes an input image and produces multi-level convolutional feature maps from shallow to deep; while in the later, to achieve accurate boundary inference and semantic segmentation, boundary-aware high resolution feature maps in shallower layers and high-level but low-resolution features are recursively embedded into the learning framework (from deep to shallow) to generate a fused feature representation that draws a holistic picture of not only high-level semantic information but also low-level fine-grained details. Experimental results on two widely-used high resolution remote sensing data sets for semantic segmentation tasks, ISPRS Potsdam and Inria Aerial Image Labeling Data Set, demonstrate competitive performance obtained by the proposed methodology compared to other studied approaches.
\end{abstract}


\end{frontmatter}


\section{Introduction}

\begin{figure*}[!t]
\centering
\includegraphics[width=0.85\linewidth]{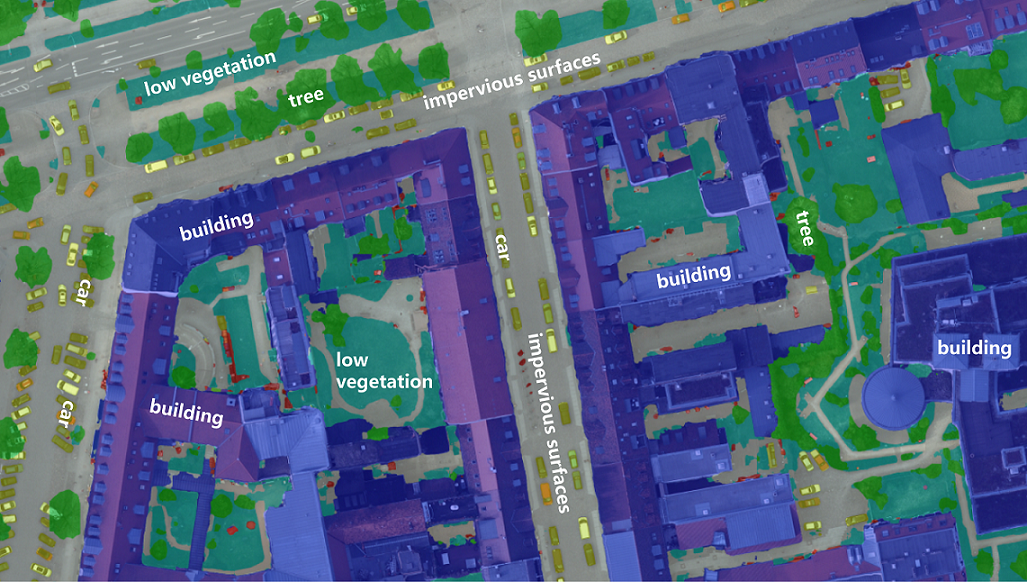}
\renewcommand{\figurename}{Fig}
\caption{\label{fig:example} An example of semantic segmentation produced with the proposed network, over a scene taken from the ISPRS Potsdam data set.}
\end{figure*}

\label{sec:intro}
Along with the launch of satellites and widespread availability of aeroplanes and unmanned aerial vehicles (UAVs), high resolution remote sensing images are now accessible at a reasonable cost. Automatic interpretation of such high resolution data is a task of primary importance for a wide range of practical applications, to name a few, land cover mapping~\cite{Marmanis17,beyongrgb,Marcos18,Maggiori17}, urban planning, and traffic monitoring~\cite{Liukang,segbeforedet}. One crucial step towards understanding a high resolution remote sensing image is to perform semantic segmentation, which consists in labeling every pixel in the image with the semantic category of the object it belongs to. The result of semantic segmentation (cf. Fig.~\ref{fig:example}) can answer the following two questions: 1) What land cover categories are observed in the image? 2) where do they appear? Semantic segmentation can basically be considered a comprehensive task that combines the traditional problems of multi-label recognition, detection, and segmentation in a single process.

\subsection{The Challenges of Semantic Segmentation for High Resolution Images}
In comparison with hyper- and multi-spectral data, images at high resolution (GSD 5-30 cm) have pretty different characteristics, bringing challenges for semantic segmentation purposes. On the one hand, intricate spatial details (e.g., roof tiles, road markings, shadows of buildings, windows of vehicles, and branches of trees) emerge, which leads to big differences in visual appearance within an object class. On the other hand, the spectral resolution of high spatial resolution sensors is usually limited to four (R-G-B-IR) or three (R-G-B) bands, so available spectral signatures are less discriminative. E.g., some roofs look quite like to roads in color channels. This is also true for low plants and trees. Hence, an effective feature representation is a matter of great importance to a semantic segmentation system for high resolution remote sensing images.

\subsection{Semantic Segmentation Using Feature Engineering}
Earlier efforts have focused on extracting useful low-level, hand-crafted visual features and/or modeling mid-level semantic features on local portions of images (e.g., patches and superpixels\footnote{A superpixel can be defined
as a set of locally connected similar pixels that preserve
detailed edge-structures for a fine segmentation.}); subsequently, a supervised classifier is employed to learn a mapping from the features to semantic categories. For example, in~\cite{Mura10}, the authors propose a morphological attribute profile (AP), which is capable of extracting effective features from spatial domain, for semantic segmentation of high resolution remote sensing images. Tokarczyk et al.~\cite{Tokarczyk15} use a boosting classifier to directly choose an optimal, comprehensive feature bank from a vast randomized quasi-exhaustive set of low-level feature candidates, in order to avoid manual feature selection.

\subsection{Semantic Segmentation Using Deep Networks}
The aforementioned methods mainly rely on manual feature engineering to build a semantic segmentation system. Recently, deep neural networks, especially convolutional neural networks (CNNs), have become the state-of-the-art model in many computer vision~\cite{Hinton12,VGG,densenet,rcnn,fastrcnn,fasterrcnn,FCN,SegNet,Moutnnls} and remote sensing problems~\cite{Zhu17DLinRS,Marcos18,DFC16,beyongrgb,Moudfc16,Marmanis17,Mou18,Maggiori17,DLinRS,Moujurse17}, as they are able to automatically extract high-level features from raw images for visual analysis tasks in an end-to-end fashion. Semantic segmentation tasks in remote sensing data are also approached by means of CNNs. Sherrah~\cite{Sherrah16} uses a fully convolutional network (FCN)~\cite{FCN} trained on natural images as a pre-trained model and fine-tunes it on high resolution remote sensing images for semantic segmentation tasks. To make use of both color image and digital surface model (DSM) data as input, while respecting their different statistical properties, Marmanis et al.~\cite{Marmanis16} employ a late fusion approach with two structurally identical, parallel FCNs. In~\cite{Kampffmeyer16}, the authors focus on small object (e.g., car) segmentation through quantifying the uncertainty at a pixel level for FCNs. By doing so, they can achieve high overall accuracy, while still achieving good accuracy for small objects. Recently, Maggiori et al.~\cite{Maggiori17} introduce a multilayer perceptron (MLP) on the top of a base FCN to learn how to effectively combine intermediate features to offer a better segmentation result. In~\cite{Audebert16}, Audebert et al. investigate the use of another network architecture, SegNet~\cite{SegNet,SegNet2}, for semantic segmentation of high resolution aerial images. In addition, they use a residual correction to perform data fusion from heterogeneous data (i.e., optical image and DSM). Later, in~\cite{beyongrgb}, they systematically study different network architectures for semantic segmentation of multimodal high resolution remote sensing data and, more specifically, they find that late fusion makes it possible to recover errors streaming from ambiguous data while early fusion allows for better joint feature learning but at the cost of higher sensitivity to missing data. Volpi and Tuia~\cite{Volpi17} compare a SegNet architecture with a standard CNN performing patch classification for semantic segmentation purposes. Marcos et al.~\cite{Marcos18} propose a segmentation network architecture called rotation equivariant vector field network (RotEqNet) to encode rotation equivariance in the network itself. By doing so, the network can be confronted with a simpler task, as it does have to learn specific weights to address rotated versions of the same object class. Marmanis et al.~\cite{Marmanis17} propose a two-step model that learns a CNN to separately output edge likelihoods at multiple scales from color-infrared (CIR) and height data. Then, the boundaries detected with each source are added as an extra channel to each source, and an FCN or SegNet is trained for semantic segmentation purposes. The intuition behind this work is that using predicted boundaries helps to achieve sharper segmentation maps.

\subsection{The motivation of This Work}
As our survey of related work shows above, most state-of-the-art CNN architectures for semantic segmentation of aerial images mainly focus on the non-linear combination of high-level features extracted from last convolutional layers. These networks, however, tend to blur object boundaries and visually degrade results due to the lack of low-level visual information that exist in shallower layers. Although, in the computer vision field, there have been some works that make an attempt to mitigate the poor localization of object boundaries either by using dilated convolution or by adding skip connections from early to deep layers of a network, they do not work well enough for remote sensing data~\cite{Marmanis17}. From the above discussions, we note that 1) how to find an effective strategy to fuse multi-level features, and 2) how to preserve object boundaries should be the most intrinsic problems in semantic segmentation of remote sensing images.
\par
To address these problems, in this paper, we propose a novel network architecture, recurrent network in fully convolutional network (RiFCN), which uses a recurrent way to fuse all-level features of a classification network (e.g., VGG-16~\cite{VGG}), while preserving boundaries as far as possible. Our work contributes to the literature in the following respects:
\begin{itemize}
  \item We propose an end-to-end trainable, bidirectional network architecture, which is composed of a forward stream and a backward stream, for the generation and fusion of multi-level convolutional features. Learning such a network architecture for pixel-wise annotations in remote sensing data has not been investigated yet to the best of our knowledge.
  \item A deep structure in the form of a recurrent network is proposed to achieve the backward stream. It embeds high-level features into low-level ones, layer by layer, and finally incorporates boundary-aware feature maps from the shallowest layer to achieve more accurate object boundary inference and semantic segmentation. The whole network architecture can be trained end-to-end by gradient learning, due to differential properties of all components.
  \item We theoretically analyze and discuss the bidirectional network learning, i.e., the backward gradient pass of the proposed network. This helps us to better understand how dose the network learn and update its weights.
\end{itemize}
\par
The paper is organized as follows. After the introductory Section~\ref{sec:intro} detailing semantic segmentation of remote sensing images, Section~\ref{sec:rw} is dedicated to a brief review of representative networks for semantic segmentation tasks in computer vision. Section~\ref{sec:med} then describes details of the proposed network. The experimental results are provided in Section~\ref{sec:exp}. Finally, Section~\ref{sec:con} concludes the paper.

\section{Representative Networks for Semantic Segmentation}
\label{sec:rw}
In this section, we would like to briefly review two representative network architectures for semantic segmentation in computer vision, namely FCN-based and encoder-decoder architecture, which both are also widely used models in semantic segmentation of remote sensing images.

\subsection{FCN-based Architecture}
Long et al.~\cite{FCN} first proposed FCN for semantic segmentation tasks, which is both efficient and effective. The key insight of FCN is that fully connected layers in a network for image classification purposes can be considered convolutions with kernels that cover their entire input region. This is equivalent to evaluating the original classification network on overlapping regions of patches. Since computations are shared across overlapping the regions, FCN is more efficient. After convolutionalizing fully connected layers in a classification network pretrained on natural images, final feature maps need to be upsampled because of the existence of pooling operations in the network. In the original FCN~\cite{FCN}, the authors enhanced output feature maps with features from intermediate layers, which is able to enable FCN to make finer predictions. Later, some extensions of FCN have been proposed to improve the performance of semantic segmentation. For example, Chen et al.~\cite{deeplab} removed some of max-pooling layers and, accordingly, introduced atrous convolutions in a FCN, which can expand the field of view without increasing the number of parameters. Furthermore, structured prediction has been studied with integrated structured models such as conditional random field (CRF). Better classification network architectures also provide new insights, e.g., ResNet\cite{ResNet}-based FCN~\cite{resfcn}. In~\cite{pspnet}, the authors proposed a pyramid pooling module and applied it to a ResNet-based network architecture. The intuition behind this model is that global parsing matters because it provides clues on the distribution of semantic categories, and the pyramid pooling module captures this information by utilizing large kernel pooling layers.

\subsection{Encoder-Decoder Architecture}
Inspired by probabilistic auto-encoders~\cite{autoencoder,resconvdeconv}, encoder-decoder paradigm has been introduced in semantic segmentation. A clear example of this branch is SegNet proposed by Badrinarayanan et al.~\cite{SegNet,SegNet2}, where the encoder is a vanilla CNN (e.g., VGG-16~\cite{VGG}) that is trained to classify images while the decoder is used to upsample the output of the encoder. The latter is composed by a set of upsampling layers and convolutional layers which are at last followed by a softmax layer to predict pixel-wise labels. Each upsampling layer in the decoder corresponds to a max-pooling layer in the encoder, and upsampled feature maps are then convolved with a set of filters to produce more dense features with finer resolution. When feature maps have been restored to a desired full resolution, they are fed to the softmax layer to produce the final segmentation map. In~\cite{DeconvNet}, the authors proposed DeconvNet, which shares a similar idea with SegNet. Moreover, U-Net~\cite{unet} can be considered an extension of SegNet, by introducing concatenations between the corresponding encoder and decoder layers. RefineNet~\cite{refinenet}, a recent network, adopts a structure similar to U-Net, but introduces several residual convolutional units in both the encoder and decoder.

\section{Methodology}
\label{sec:med}
\subsection{An Observation}

\begin{figure*}[!t]
\centering
\includegraphics[width=\linewidth]{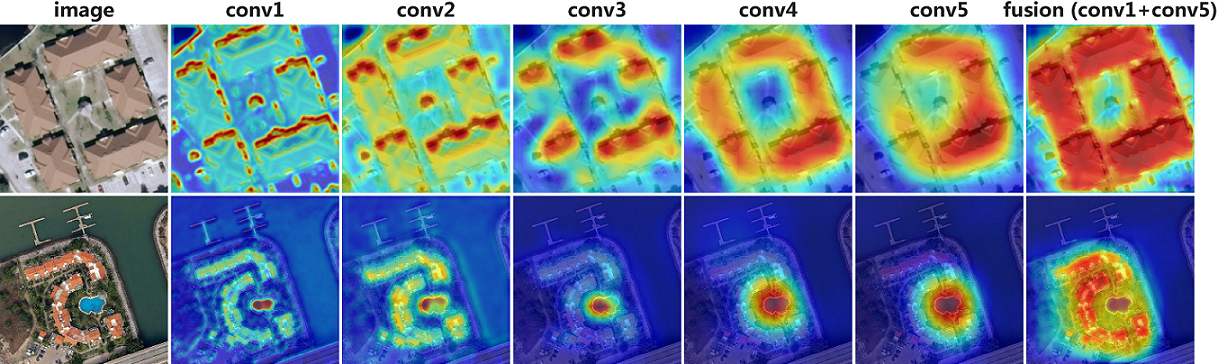}
\renewcommand{\figurename}{Fig}
\caption{\label{fig:cams} Feature responses from different convolutional stages of a VGG model, in which class activation maps (CAMs) generated from shallower layers present an explicit view of low-level features and CAMs from deeper layers highlight coarse discriminative regions. The CAMs generated from the fused features using~\cite{Hua18} draw a holistic picture of not only where discriminative regions are, but also how the regions appear in detail.}
\end{figure*}

Recently, several studies~\cite{Zeiler14,Mahendran15} that attempt to reveal what learned by CNNs using a gradient guided technique show that deeper layers make use of filters to grasp global high-level information while shallower layers capture local low-level details such as object boundaries and edges. A work in this direction for remote sensing images can be found in~\cite{Hua18}, where the authors make use of class activation maps (CAMs)~\cite{CAM} to visualize learned feature maps of a CNN and come to an almost same conclusion, i.e., CAMs generated from shallower layers present an explicit view of low-level features, and CAMs from deeper layers highlight coarse discriminative regions. In addition, they design a network that only fuses feature maps of the first and the last stage, and such a network can achieve a significant improvement in terms of classification accuracy. The CAM generated from the fused features draw a holistic picture of not only where discriminative regions are, but also how the regions appear in detail (cf. Fig.~\ref{fig:cams}). This work gives us an incentive to design a network that is capable of exploiting all the features available along the forward process of a CNN to generate a holistic feature for semantic segmentation of aerial images. In this way, shallower layers that capture fine-grained features can be directly refined using high-level semantic features from deeper layers. To this end, we need to come up with a solution that can sequently and progressively embed high-level features into low-level features. Recurrent neural networks (RNNs) have gained significant attention for solving many challenging problems involving sequential data analysis and recently shown to be successful in several remote sensing applications~\cite{Lyu16,Mournn,Russwurm17,recnn,Russwurm18,Lyu18}. Therefore, in this work, we would like to make use of the idea of RNNs to achieve the sequential fusion of all-level features in our network.

\subsection{Network Architecture}
\begin{figure*}[!t]
\centering
\includegraphics[width=\linewidth]{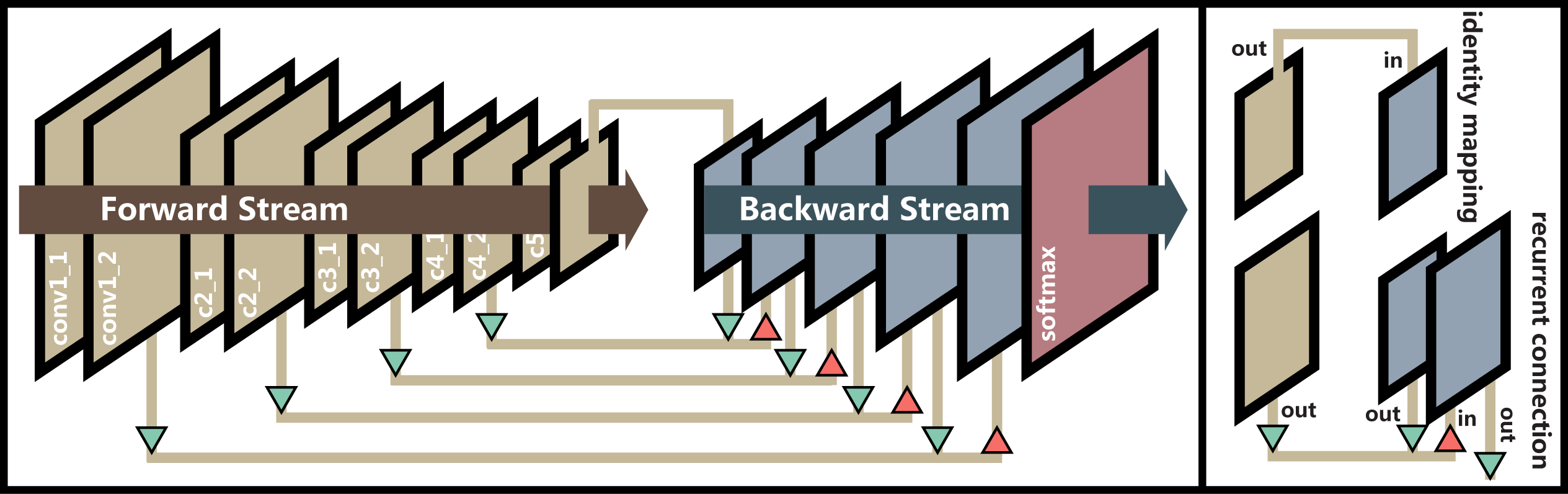}
\renewcommand{\figurename}{Fig}
\caption{\label{fig:net} Overall architecture of the proposed recurrent network in fully convolutional network (RiFCN) for semantic segmentation of aerial images. RiFCN is a bidirectional network, which has a forward stream and a backward stream, followed by a final pixel-wise classification layer. The forward stream is a CNN for feature extraction, which takes an input image and produces multi-level convolutional feature maps; while the backward stream incorporates autoregressive recurrent connections to hierarchically and progressively absorb abstract, high-level features and render pixel-wise, high resolution prediction. The latter can be considered a reverse feature fusion process.}
\end{figure*}

The proposed bidirectional network, RiFCN, has a forward stream and a backward stream, followed by a final pixel-wise classification layer. The forward stream is a CNN for feature extraction, which takes an input image and produces multi-level convolutional feature maps; while the backward stream exploits all the features available along the forward stream to enable high resolution prediction using recurrent connections. Fig.~\ref{fig:net} shows the overall architecture of the proposed RiFCN.

\par
\subsubsection{Forward Stream}
The forward stream in our network is mainly inspired by the philosophy of VGG-16~\cite{VGG}, which is well known for its elegance and simplicity and, at the same time, yields nearly state-of-the-art features for image classification and good generalization properties. More specifically, the forward stream consists of 5 convolutional blocks (2 convolutional layers per block). Note that we do not initialize the training process of the network from weights trained for classification on large natural image data set like ImageNet, as the pre-trained model is not suitable to be used on multi-channel images (e.g., R-G-B-IR). In addition, there are no fully connected layers, in order to significantly reduce the number of trainable parameters in the network and retain higher resolution feature maps at the same time.
\par
We make use of convolutional filters with a very small receptive field of $3\times3$, rather than using larger ones, such as $5\times5$ or $7\times7$. That is because, as reported in~\cite{VGG}, $3\times3$ convolutional filters are the smallest kernels that can capture patterns in various directions (e.g., center, up/down, and left/right), increase nonlinearities inside the network, and thus make the network more discriminative as compared to other larger filters. The spatial padding of convolutional layers is such that the spatial resolution of feature maps is preserved after convolution, i.e., it is 1 pixel in our network; the convolution stride is fixed to 1 pixel. Spatial pooling is achieved by carrying out 4 max-pooling layers, which follow the first four convolutional blocks. Max-pooling is performed over $2\times2$ pixel window with stride 2.
\par
In the forward stream of RiFCN, convolutional layers within the same block have the same number of filters. Meanwhile, the number of filters increases in the deeper blocks, roughly doubling after each max-pooling layer, which is meant to preserve time complexity per layer as far as possible. We make use of 64 filters for the first two convolutional layers, 128 filters for the following two layers, 256 filters for the fifth and the sixth convolutional layer, 512 filters for the seventh and the eighth layer, and 1024 filters for the last two convolutional layers. All the convolutional layers in the forward stream have a rectified linear unit (ReLU) nonlinearity. In addition, note that the forward stream in our network is extremely flexible in that it can be replaced and modified in various ways, for example, using other classification network architectures (VGG-19~\cite{VGG}, ResNet~\cite{ResNet}, etc.).

\begin{figure*}[!t]
\centering
\includegraphics[width=\linewidth]{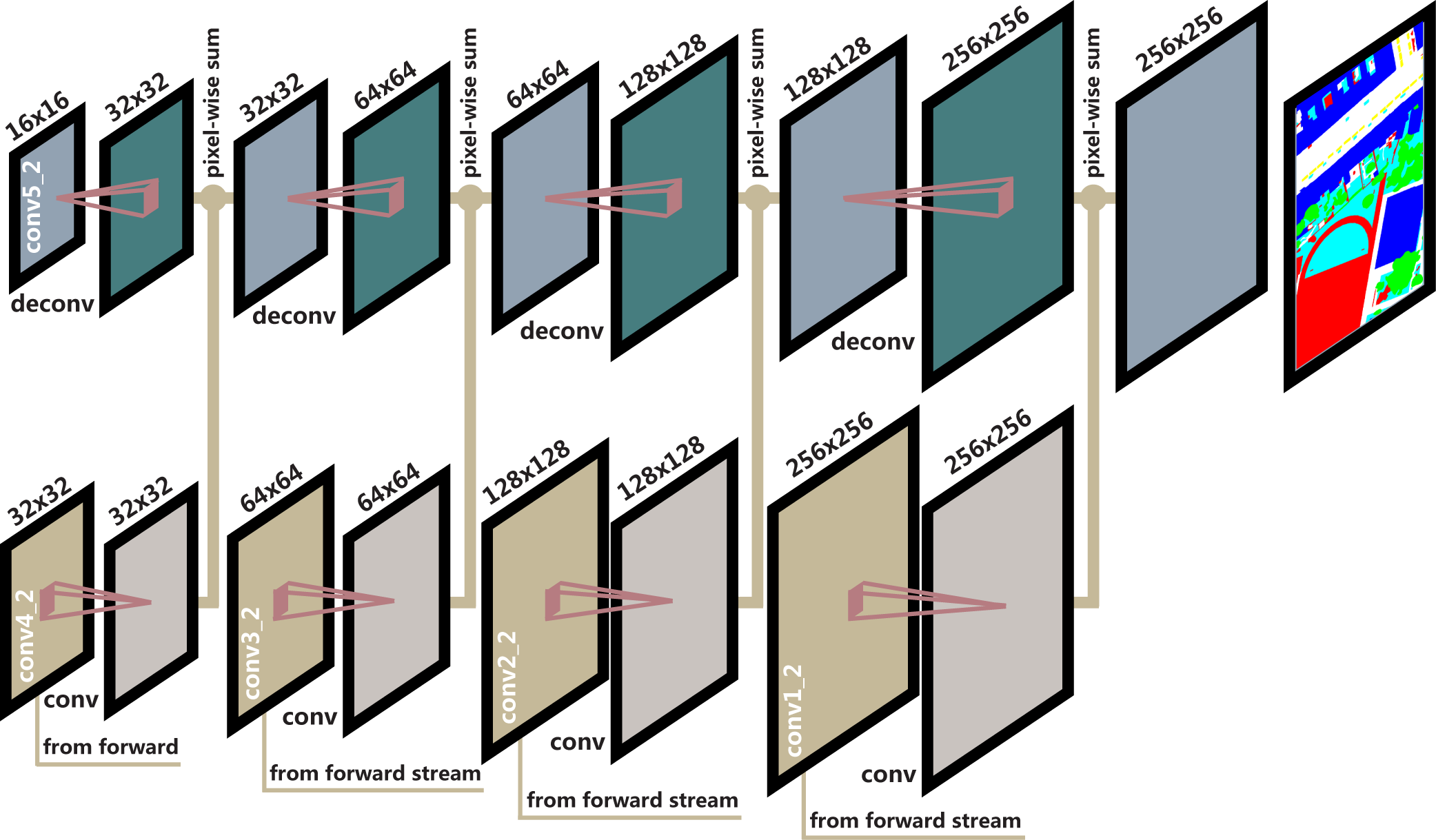}
\renewcommand{\figurename}{Fig}
\caption{\label{fig:recurrent} Details of the backward stream of RiFCN based on autoregressive recurrent connection. During the backward pass, in each level, it takes feature maps generated by the forward pass and fused feature maps from the previous level as inputs to produce new features of the current level.}
\end{figure*}

\subsubsection{Backward Stream}
The forward stream is used to extract different-level features of images, by interleaving convolutional layers and max-pooling layers, i.e., spatially shrinking the feature maps. Pooling is necessary to allow the aggregation of information over large areas of feature maps and, more fundamentally, to make network training computationally feasible. It, however, leads to reduced resolutions of high-level feature maps in deeper layers. Therefore, to provide dense pixel-wise predictions, we need a way to refine those coarse, pooled feature maps.
\par
A straightforward idea is to deconvolve all the feature maps to the desired full resolution and stack them together, resulting in a concatenated feature representation that can be used to predict segmentation maps~\cite{FCN}. Although this kind of approaches are capable of semantically segment images from different levels, the inner connection of different-level features is missing. Furthermore, in this approach, fusing earlier layers easily result in diminishing returns, with respect to both visual and quantitative improvements~\cite{FCN}. Another way~\cite{SegNet} is using a deconvolutional network which takes as input only coarse, high-level feature maps. This approach does not exploit low-level features that help to generate sharp, detailed boundaries for high resolution prediction.
\par
\begin{figure*}[!t]
\centering
\includegraphics[width=0.7\linewidth]{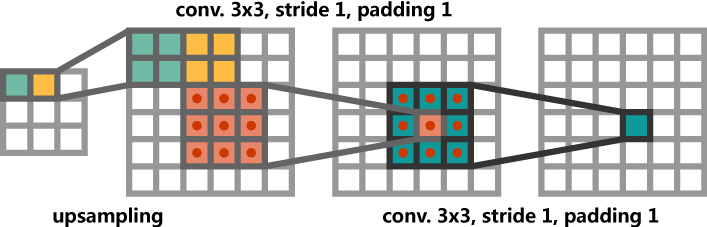}
\renewcommand{\figurename}{Fig}
\caption{\label{fig:deconv} Deconvolution used in the recurrent connection of the backward stream.}
\end{figure*}

In this work, we propose a novel feature fusion architecture based on a recurrent structure to hierarchically and progressively absorb high-level semantic features and render pixel-wise, high resolution predictions. It incorporates autoregressive recurrent connections into predictions from deep to shallow layers, which is opposite to the forward stream. Fig.~\ref{fig:recurrent} shows the detailed process of the backward stream in our RiFCN. Formally, given an input image with size $W\times H$, output features maps of the forward stream have size $[\frac{W}{2^4},\frac{H}{2^4}]$, i.e., the output features are reduced by a factor of 16. Let $\kappa=[\frac{W}{2^l},\frac{H}{2^l}]$ be the resolution of feature maps and identified by feature level $l(=0,1,\cdots,L)$; $\bm{F}_{fwd}^l$ denotes a 3D tensor, i.e., feature maps generated by the $l$-th convolutional block of the forward stream. During the backward stream, in each level $l$, it takes $\bm{F}_{fwd}^l$ and fused feature maps from the previous level $\bm{F}_{bwd}^{l+1}$ as inputs to produce new features of the current level as
\begin{equation}\label{eq:r1}
\bm{F}_{bwd}^l=
\begin{cases}
\varphi(\bm{F}_{fwd}^l,\bm{F}_{bwd}^{l+1}) &\mbox{if $l<L$}\\
\bm{F}_{fwd}^l &\mbox{if $l=L$}
\end{cases}\,,
\end{equation}
where $\varphi$ is a function for fusing different feature maps at different resolutions. We define it as follows:
\begin{equation}\label{eq:r2}
\varphi(\bm{F}_{fwd}^l,\bm{F}_{bwd}^{l+1}) = \sigma(\bm{W}_{fwd}\ast\bm{F}_{fwd}^l)+\sigma(\bm{W}_{bwd}\star_{s}\bm{F}_{bwd}^{l+1})\,,
\end{equation}
where $\ast$ represents convolution operation, and $\star_{s}$ denotes deconvolution operation with stride $s$ (cf. Fig.~\ref{fig:deconv}). $\bm{W}_{fwd}$ and $\bm{W}_{bwd}$ are weights of the convolution and deconvolution, respectively. $\sigma$ is a nonlinear function and in this work, we make use of ReLU. Note that Eq.~\ref{eq:r2} is a general expression, and the first term can be $\bm{F}_{fwd}^l$. In this case, $\bm{W}_{fwd}$ is a matrix that its center element is 1 and others are 0.
\par
From Eq.~(\ref{eq:r2}), we can clearly see that multiple autoregressive recurrent connections ensure that final fused feature maps $\bm{F}_{bwd}^{0}$ have multiple paths from deep to shallow layers, which facilitates effective information exchanges. In addition, this top-down backward stream is able to propagate semantic information back to fine-grained details for the final segmentation prediction.

\subsection{Bidirectional Network Learning}
Denote by $\{(\bm{X}_i,\bm{Y}_i)\}_{i=1}^N$ with $N$ sample pairs representing training set, where $\bm{X}_i=\{x_i^k\}$ and $\bm{Y}_i=\{y_i^k\}$ $(k=1,2,\cdots,K)$ are the input image and the corresponding ground-truth with $K$ pixels, respectively. For notional simplicity, we subsequently drop the subscript $i$ and consider each image independently. We denote $\bm{W}$ as parameters of the forward stream. Thus, the loss function of RiFCN can be expressed as
\begin{equation}\label{eq:loss}
\mathcal{L}=-\sum_{c}\sum_{k\in Y_c}\log {\rm Pr}(y^k=c|\bm{X};\bm{W},\bm{W}_{fwd},\bm{W}_{bwd})\,,
\end{equation}
where $c=1,\cdots,M$ and $M$ is the number of classes. $Y_c$ represents the $c$-th label set. ${\rm Pr}(y^k=c|\bm{X};\bm{W},\bm{W}_{fwd},\bm{W}_{bwd})\in [0,1]$ is the confidence score of the prediction that measures how likely the pixel $x^k$ belongs to the $c$-th class. Note that $\bm{W}_{fwd}$ and $\bm{W}_{bwd}$ are both parameters in the backward stream. Since network learning in the forward stream (i.e., the updating of $\bm{W}$) is similar to that of a CNN for classification tasks, in this section, we mainly focus on the network inference of the backward stream.
\par
The backward stream of the network starts with computing the gradient of the loss function $\mathcal{L}$ with respect to the output $\frac{\partial \mathcal{L}}{\partial y}$. This gradient is then propagated backwards level by level from output to input to update the parameters of the network. The recurrent structure to propagate this gradient through the network can be written as follows:
\begin{equation}\label{eq:ana1}
\frac{\partial \mathcal{L}}{\partial \bm{F}_{bwd}^{l+1}}=\frac{\partial \mathcal{L}}{\partial \bm{F}_{bwd}^{l}}\frac{\partial \bm{F}_{bwd}^{l}}{\partial \bm{F}_{bwd}^{l+1}}=\frac{\partial \mathcal{L}}{\partial \bm{F}_{bwd}^{l}}\bm{w}_{bwd}\,,
\end{equation}
and starts at:
\begin{equation}\label{eq:ana2}
\frac{\partial \mathcal{L}}{\partial y}=\frac{\partial \mathcal{L}}{\partial \bm{F}_{bwd}^{0}}\,.
\end{equation}
\par
From Eq.~\ref{eq:ana1}, we can see that only parameters that connect $\bm{w}_{bwd}$ play a role in propagating the error down the network.
\par
The gradients of the loss function with respect to the parameters can then be obtained by summing the parameter gradients in each level (or accumulating them while propagating the error):
\begin{equation}\label{eq:ana3}
\begin{split}
\frac{\partial \mathcal{L}}{\partial \bm{w}_{bwd}}&=\sum_{l=0}^{L}\frac{\partial \mathcal{L}}{\partial \bm{F}_{bwd}^{l}}\bm{F}_{bwd}^{l+1}\\
\frac{\partial \mathcal{L}}{\partial \bm{w}_{fwd}}&=\sum_{l=0}^{L}\frac{\partial \mathcal{L}}{\partial \bm{F}_{bwd}^{l}}\bm{F}_{fwd}^{l}\,.
\end{split}
\end{equation}
\par
The momentum method is commonly used to help accelerate stochastic gradient descent in the relevant direction and dampen oscillations by adding a fraction $\gamma$ of the update parameter. When updating weights $\bm{w}_{bwd}$ and $\bm{w}_{fwd}$ using the momentum method, the updating rules can be written as follows:
\begin{equation}\label{eq:ana3}
\begin{split}
{\Delta \bm{w}_{bwd}}&:=\gamma{\Delta \bm{w}_{bwd}}+\eta\sum_{l=0}^{L}\frac{\partial \mathcal{L}}{\partial \bm{F}_{bwd}^{l}}\bm{F}_{bwd}^{l+1}\\
{\Delta \bm{w}_{fwd}}&:=\gamma{\Delta \bm{w}_{fwd}}+\eta\sum_{l=0}^{L}\frac{\partial \mathcal{L}}{\partial \bm{F}_{bwd}^{l}}\bm{F}_{fwd}^{l}\,,
\end{split}
\end{equation}
where $\eta$ is the learning rate and $\gamma$ is the momentum.

\begin{figure*}[!t]
\centering
\includegraphics[width=\linewidth]{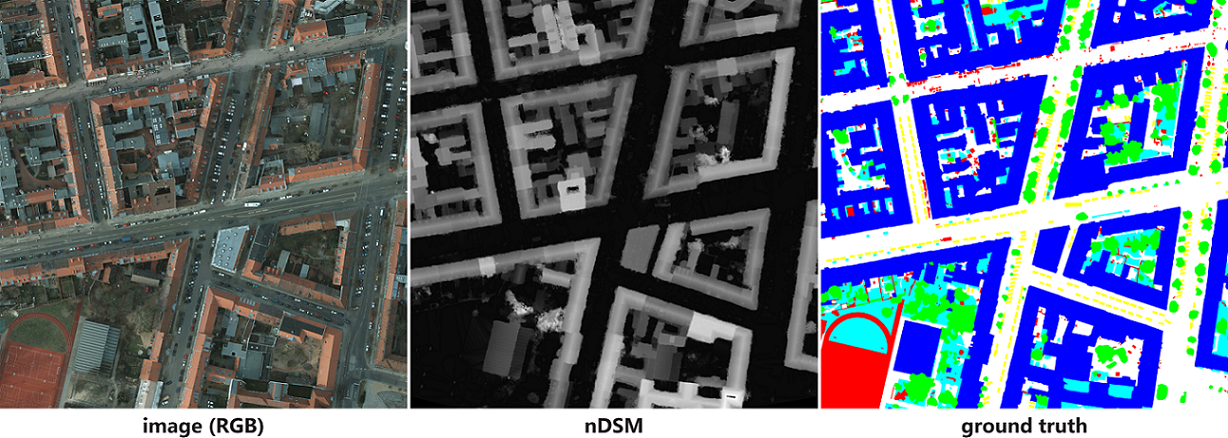}
\renewcommand{\figurename}{Fig}
\caption{\label{fig:isprs_data} An excerpt from the ISPRS Potsdam data set for semantic segmentation. Legend -- white: impervious surfaces, \textcolor[rgb]{0,0,1}{blue}: buildings, \textcolor[rgb]{0,1,1}{cyan}: low vegetation, \textcolor[rgb]{0,1,0}{green}: trees, \textcolor[rgb]{1,1,0}{yellow}: cars, \textcolor[rgb]{1,0,0}{red}: clutter/background.}
\end{figure*}

\begin{figure}[!t]
\centering
\includegraphics[width=\columnwidth]{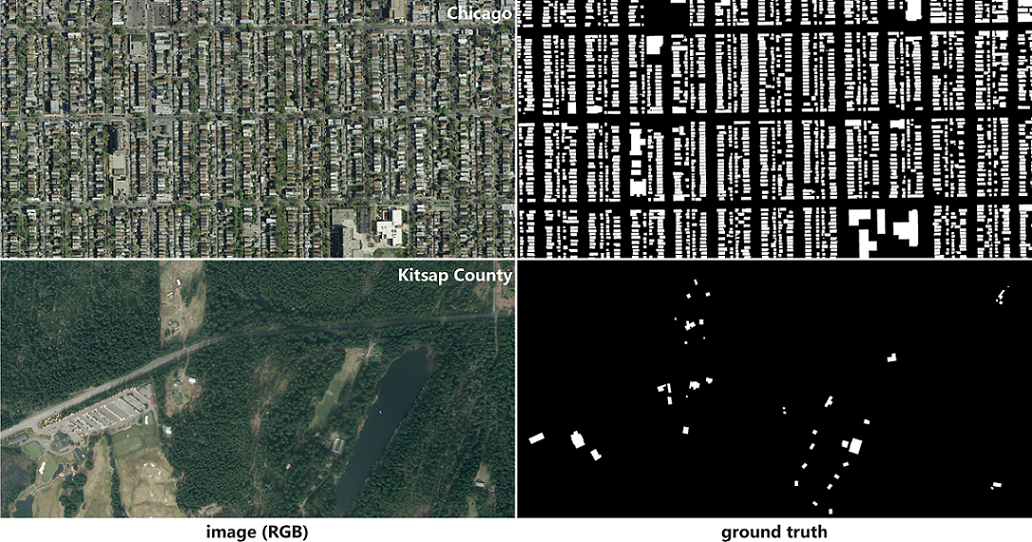}
\renewcommand{\figurename}{Fig}
\caption{\label{fig:inria_data} Two tiles (left) and their corresponding ground truths (right) from the Inria Aerial Image Labeling data set for semantic segmentation. The images in this data set convey dissimilar urban settlements, ranging from densely populated areas to alpine towns.}
\end{figure}


\section{Experiments}
\label{sec:exp}
\subsection{Data Sets}
\subsubsection{ISPRS Potsdam}
The ISPRS Potsdam Semantic Labeling data set is an open benchmark data set provided online\footnote{\url{http://www2.isprs.org/commissions/comm3/wg4/2d-sem-label-potsdam.html}} for semantic segmentation of high resolution remote sensing images. The dataset is consists of 38 ortho-rectified aerial IRRGB images ($6000\times6000$ px), with a 5 cm spatial resolution and corresponding DSMs generated by dense image matching, taken over the city of Potsdam, Germany. A comprehensive manually annotated pixel-wise segmentation mask is provided as ground truth for 24 tiles, which are the tiles we work on. We randomly selected 6 tiles (tile IDs: 2\_11, 3\_11, 4\_12, 5\_12, 7\_10, 7\_12) from 24 training images and used them as test set in our experiments. The input to the networks contains both IRRG and nDSM, and all results reported on this dataset refer to the aforementioned test set. Fig.~\ref{fig:isprs_data} shows an excerpt from the ISPRS Potsdam data set.

\subsubsection{Inria Aerial Image Labeling Data Set}
The Inria Aerial Image Labeling Data Set has been recently proposed and is specially designed for advancing technologies in automatic pixel-wise labeling of aerial imagery. It is comprised of 360 ortho-rectified aerial RGB images ($5000\times5000$ px) at 30 cm spatial resolution. Each tile covers a surface of $1500\times 1500$ m$^2$. All the images cover ten cities and an overall area of $810$ km$^2$. The images convey dissimilar urban settlements, ranging from densely populated areas (e.g., Chicago, USA) to alpine towns (e.g., Tyrol, Austrian). Manually annotated ground truth is only provided for five cities, namely Austin, Chicago, Kitsap County, Western Tyrol, and Vienna. The ground truth is binary, which indicates a pixel belongs either to building or non-building class. For comparability, as suggested by the authors of this data set, we use images 6 to 36 of each city for training and images 1 to 5 for testing. Two examples from the Inria Aerial Image Labeling Data Set are exhibited in Fig.~\ref{fig:inria_data}.

\subsection{Network Training}
The network training is based on the TensorFlow framework. We chose Nesterov Adam~\cite{nadam2,nadam1} as the optimizer to train the network since, for this task, it shows much faster convergence than standard stochastic gradient descent (SGD) with momentum~\cite{sgd} or Adam~\cite{adam}. We fixed almost all of parameters of Nesterov Aadam as recommended in~\cite{nadam2}: $\beta_1=0.9$, $\beta_2=0.999$, $\epsilon=1\mathrm{e}{-08}$, and a schedule decay of 0.004, making use of a fairly small learning rate of $2\mathrm{e}{-04}$. All network weights are initialized with a Glorot uniform initializer~\cite{Glorot_normal} that draws samples from a uniform distribution. We utilize softmax and sigmoid as the activation function of the last convolutional layer for multi-class and binary semantic segmentation, respectively. We make use of data augmentation technique to increase the number of training samples. The patches and corresponding pixel-wise ground truth are transformed by horizontally and vertically flipping three-quarters of the patches. We train the network for 30 epochs and use early stopping to avoid overfitting. To monitor overfitting during training, we randomly select 10\% of the training samples as the validation set. Furthermore, we exploit fairly small mini-batches of 8 image pairs because, in a sense, every pixel is a training sample. Finally, we train our network on a single NVIDIA GeForce GTX TITAN with 12 GB of GPU memory.

\begin{table}[t]
\caption{\label{tab:isprs} Numerical Results on the ISPRS Potsdam Data Set.}
\centering
\begin{adjustwidth}{-2cm}{0cm}
\begin{threeparttable}
\begin{tabular}{cccccccccccc}
\toprule[1pt]
\textbf{Method} & \textbf{Imp Surf} & \textbf{Building} & \textbf{Low Veg} & \textbf{Tree} & \textbf{Car} & \textbf{Clutter} & \textbf{OA} & \textbf{Mean $\bm{F_1}$} \\
\toprule[0.5pt]
FCN & 88.46 & \textbf{92.28} & 78.33 & 73.10 & 82.83 & 69.55 & 84.39 & 80.76 \\
SegNet & 88.53 & 91.90 & 79.68 & 76.04 & 86.51 & 61.16 & 84.68 & 80.64 \\
RiFCN & \textbf{90.10} & 92.23 & \textbf{81.94} & \textbf{79.29} & \textbf{88.91} & \textbf{69.71} & \textbf{86.59} & \textbf{83.70} \\
\toprule[0.5pt]
\toprule[0.5pt]
FCN [e] & 90.32 & \textbf{93.16} & 80.03 & 75.78 & 89.26 & \textbf{72.23} & 86.26 & 83.46 \\
SegNet [e] & 90.41 & 92.77 & 81.65 & 78.77 & 92.41 & 63.61 & 86.58 & 83.27 \\
RiFCN [e] & \textbf{91.74} & 93.02 & \textbf{83.71} & \textbf{81.90} & \textbf{93.73} & 72.18 & \textbf{88.30} & \textbf{86.05} \\
\bottomrule[1pt]
\end{tabular}
\begin{tablenotes}
\item[] [e] means evaluation on eroded boundary ground truths.
\end{tablenotes}
\end{threeparttable}
\end{adjustwidth}
\end{table}

\begin{figure*}[h]
\centering
\includegraphics[width=\linewidth]{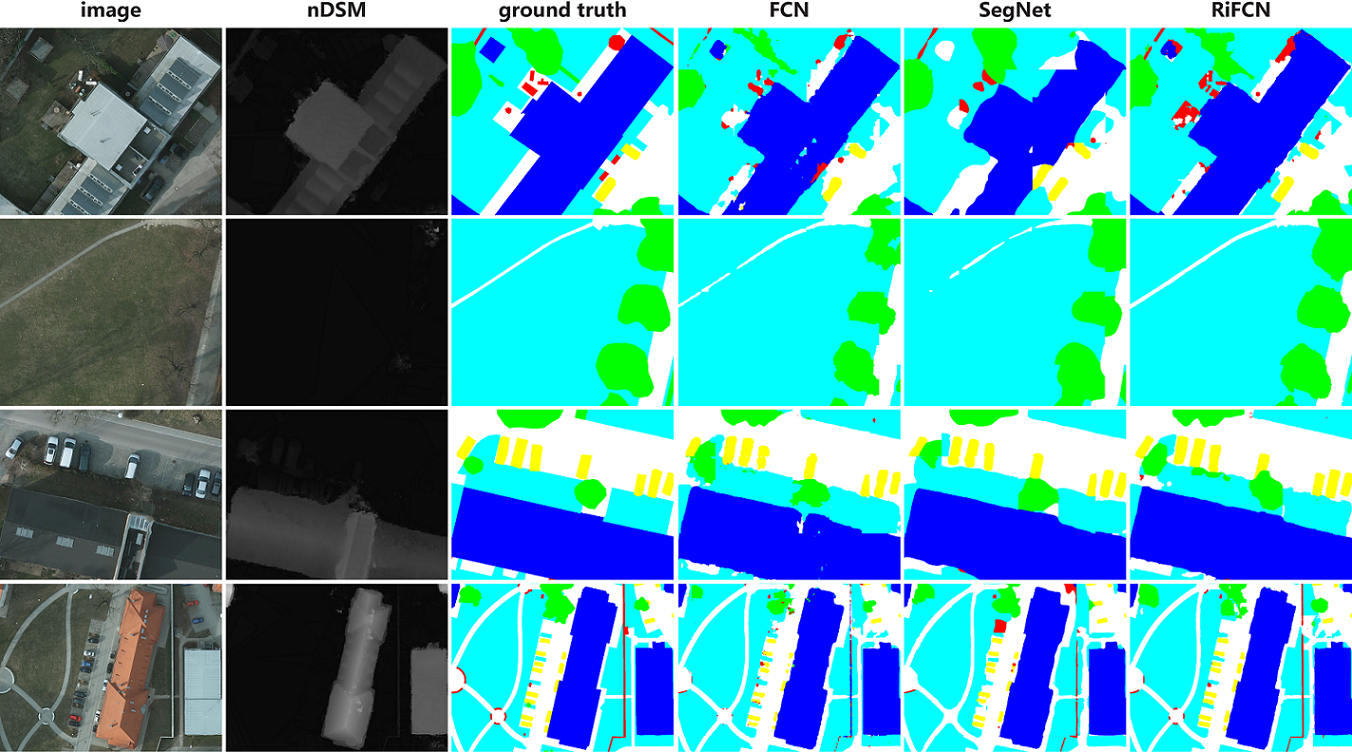}
\renewcommand{\figurename}{Fig}
\caption{\label{fig:isprs_zoom} Example predictions of different models on the ISPRS Potsdam data set. Legend -- white: impervious surfaces, \textcolor[rgb]{0,0,1}{blue}: buildings, \textcolor[rgb]{0,1,1}{cyan}: low vegetation, \textcolor[rgb]{0,1,0}{green}: trees, \textcolor[rgb]{1,1,0}{yellow}: cars, \textcolor[rgb]{1,0,0}{red}: clutter/background.}
\end{figure*}

\subsection{ISPRS Potsdam Results}
To evaluate the performance of different methods for semantic segmentation of aerial images, $F_1$ score and overall accuracy are used as evaluation criteria. $F_1$ score can be calculated as follows:
\begin{equation}\label{eq:eva1}
F_1=2\times\frac{precision\times recall}{precision+recall}\,,
\end{equation}
and
\begin{equation}\label{eq:eva2}
precision=\frac{tp}{tp+fp}, \quad recall=\frac{tp}{tp+fn}\,,
\end{equation}
where $tp$, $fp$, and $fn$ are the numbers of true positives, false positives, and false negatives, respectively. These metrics can be calculated by pixel-based confusion matrices per tile, or an accumulated confusion matrix. Overall accuracy is the normalization of the trace from the confusion matrix.

\begin{figure*}[!t]
\centering
\includegraphics[width=\linewidth]{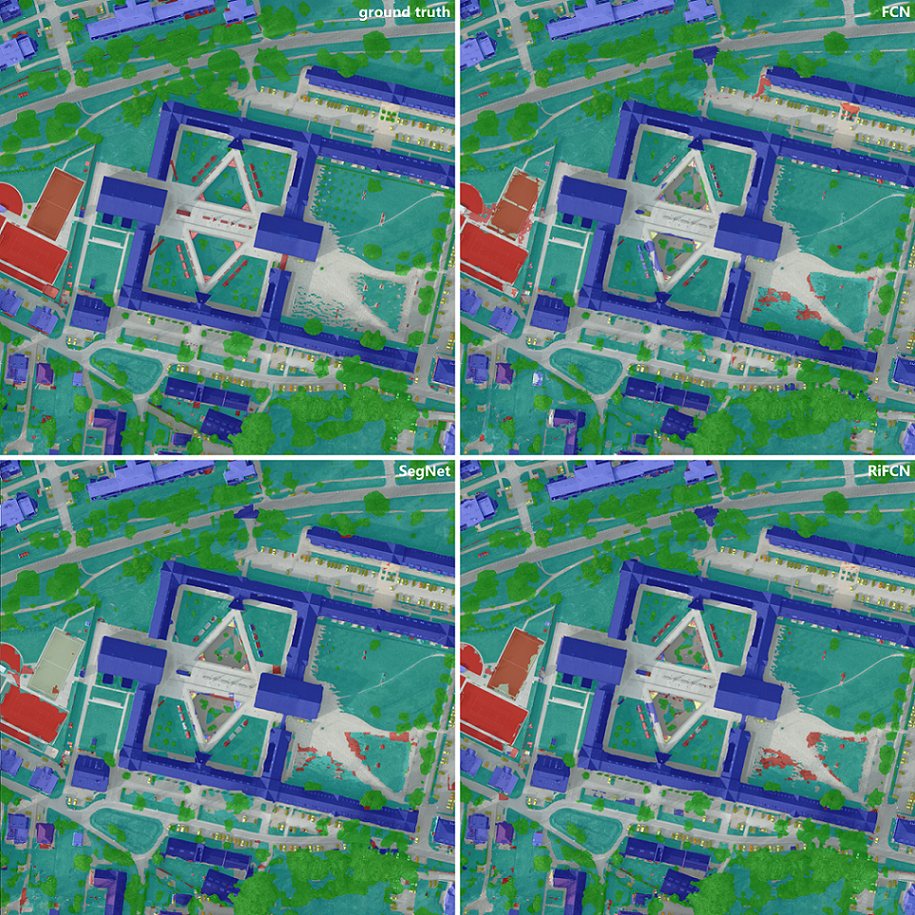}
\renewcommand{\figurename}{Fig}
\caption{\label{fig:isprs_full} Full prediction for tile ID 3\_11. The mean $F_1$ scores achieved by FCN, SegNet, and RiFCN are 82.49\%, 80.36\%, and 87.31\%, respectively; the overall accuracies of these methods on this tile are 85.07\%, 85.14\%, and 87.31\%, respectively. Legend -- white: impervious surfaces, \textcolor[rgb]{0,0,1}{blue}: buildings, \textcolor[rgb]{0,1,1}{cyan}: low vegetation, \textcolor[rgb]{0,1,0}{green}: trees, \textcolor[rgb]{1,1,0}{yellow}: cars, \textcolor[rgb]{1,0,0}{red}: clutter/background.}
\end{figure*}

To verify the effectiveness of the proposed network, we perform comparisons against a couple of state-of-the-art semantic segmentation networks, i.e., FCN and SegNet, which are two most widely used models in semantic segmentation of aerial images. Note that we do not compare RiFCN with other networks in computer vision (e.g., PSPNet~\cite{pspnet}), as they make use of some techniques (fully connected CRF, ResNet, etc.) which leads to an unfair comparison. Table~\ref{tab:isprs} presents results on the ISPRS Potsdam data set, and we can see that RiFCN significantly outperforms other methods on both $F_1$ score and overall accuracy. Compared to FCN, the proposed RiFCN increases the mean $F_1$ score and overall accuracy by 2.94\% and 2.20\%, respectively; in comparison with SegNet, increments on mean $F_1$ score and overall accuracy are 3.06\% and 1.91\%, respectively. Moreover, it is worth noting that the proposed network can achieve the best accuracy on small objects (e.g., cars). The comparisons indicate that the good performance of RiFCN can be ascribed to the proposed top-down backward stream, which effectively fuses multi-level features using autoregressive recurrent connections. Some semantic segmentation results on the ISPRS Potsdam data set are shown in Fig.~\ref{fig:isprs_zoom}. We can see an improvement in visual quality from FCN and SegNet to RiFCN.

\begin{figure*}[!t]
\centering
\includegraphics[width=\linewidth]{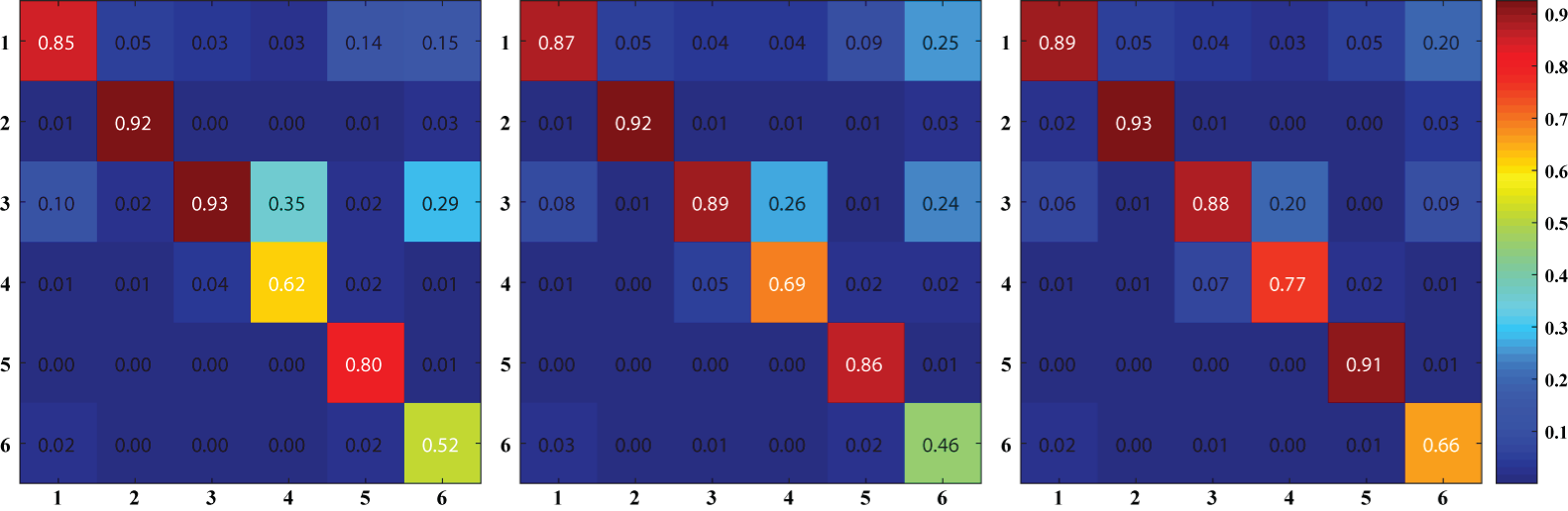}
\renewcommand{\figurename}{Fig}
\caption{\label{fig:CMs} Confusion matrices of FCN (left), SegNet (middle), and RiFCN (right) for the ISPRS Potsdam data set.}
\end{figure*}

In addition, according to the evaluation of the ISPRS Potsdam data set, we also report results based on an alternative ground truth (cf.~the last three rows in Table~\ref{tab:isprs}) in which the borders have been eroded by a 3 px radius circle, so that evaluation is tolerant to small errors on object edges. In Fig.~\ref{fig:isprs_full}, the full segmentation of
image tile 3\_11 is given. It summarizes the classification of an entire tile.
\par
Confusion matrices of different networks for the ISPRS Potsdam data set can be found in Fig.~\ref{fig:CMs}. We can clearly see that the proposed RiFCN can better differentiate similar classes as compared to FCN and SegNet.

\begin{table}[t]
\caption{\label{tab:inria} Numerical Results on the Inria Aerial Image Labeling Data Set.}
\centering
\begin{adjustwidth}{-2.95cm}{0cm}
\begin{threeparttable}
\begin{tabular}{cccccccccccc}
\toprule[1pt]
\textbf{Method} &  & \textbf{Austin} & \textbf{Chicago} & \textbf{Kitsap County} & \textbf{Western Tyrol} & \textbf{Vienna} & \textbf{Overall} \\
\toprule[0.5pt]
\multirow{2}{*}{FCN} & IoU & 47.66 & 53.62 & 33.70 & 46.86 & 60.60 & 53.82 \\
 & Acc. & 92.22 & 88.59 & 98.58 & 95.83 & 88.72 & 92.79 \\
\toprule[0.5pt]
\multirow{2}{*}{FCN-Skip} & IoU & 57.87 & 61.13 & 46.43 & 54.91 & 70.51 & 62.97 \\
 & Acc. & 93.85 & 90.54 & 98.84 & 96.47 & 91.48 & 94.24 \\
\toprule[0.5pt]
\multirow{2}{*}{FCN-MLP} & IoU & 61.20 & 61.30 & 51.50 & 57.95 & 72.13 & 64.67 \\
 & Acc. & 94.20 & 90.43 & 98.92 & 96.66 & 91.87 & 94.42 \\
\toprule[0.5pt]
\toprule[0.5pt]
\multirow{2}{*}{SegNet} & IoU & 74.81 & 52.83 & 68.06 & 65.68 & 72.90 & 70.14 \\
 & Acc. & 92.52 & 98.65 & 97.28 & 91.36 & 96.04 & 95.17 \\
\toprule[0.5pt]
\multirow{2}{*}{Multi-task SegNet$^*$} & IoU & 76.76 & 67.06 & \textbf{73.30} & 66.91 & 76.68 & 73.00 \\
 & Acc. & 93.21 & \textbf{99.25} & 97.84 & 91.71 & \textbf{96.61} & 95.73 \\
\toprule[0.5pt]
\toprule[0.5pt]
\multirow{2}{*}{Mask R-CNN} & IoU & 65.63 & 48.07 & 54.38 & 70.84 & 64.40 & 59.53 \\
 & Acc. & 94.09 & 85.56 & 97.32 & \textbf{98.14} & 87.40 & 92.49 \\
\toprule[0.5pt]
\toprule[0.5pt]
\multirow{2}{*}{RiFCN} & IoU & \textbf{76.84} & \textbf{67.45} & 63.95 & \textbf{73.19} & \textbf{79.18} & \textbf{74.00} \\
 & Acc. & \textbf{96.50} & 91.76 & \textbf{99.14} & 97.75 & 93.95 & \textbf{95.82} \\
\bottomrule[1pt]
\end{tabular}
\begin{tablenotes}
\item[] $^*$ This method uses extra supervision information for network training.
\end{tablenotes}
\end{threeparttable}
\end{adjustwidth}
\end{table}

\subsection{Inria Aerial Image Labeling Data Set Results}
To quantify performance, two evaluation measures are considered in this data set: intersection over union (IoU) and overall accuracy. IoU, also known as Jaccard index, is defined as follows:
\begin{equation}\label{eq:eva3}
IoU(A,B)=\frac{area(A\cap B)}{area(A\cup B)}\,.
\end{equation}
\par
The IoU is a measure of how close two regions are each other on a scale between 0 and 1 -- a value of 0 means the regions do not overlap and a scale of 1 means that the regions are exactly the same. We mainly focus experimental results on IoU as it has become a standard evaluation criterion for binary semantic segmentation tasks. In addition, given that category distribution suffers from an imbalanced phenomenon (a large number of image areas are dedicated to background/non-building class), overall accuracy is not precise enough, since building category is easy to be ignored.

\begin{figure*}[!t]
\centering
\includegraphics[width=\linewidth]{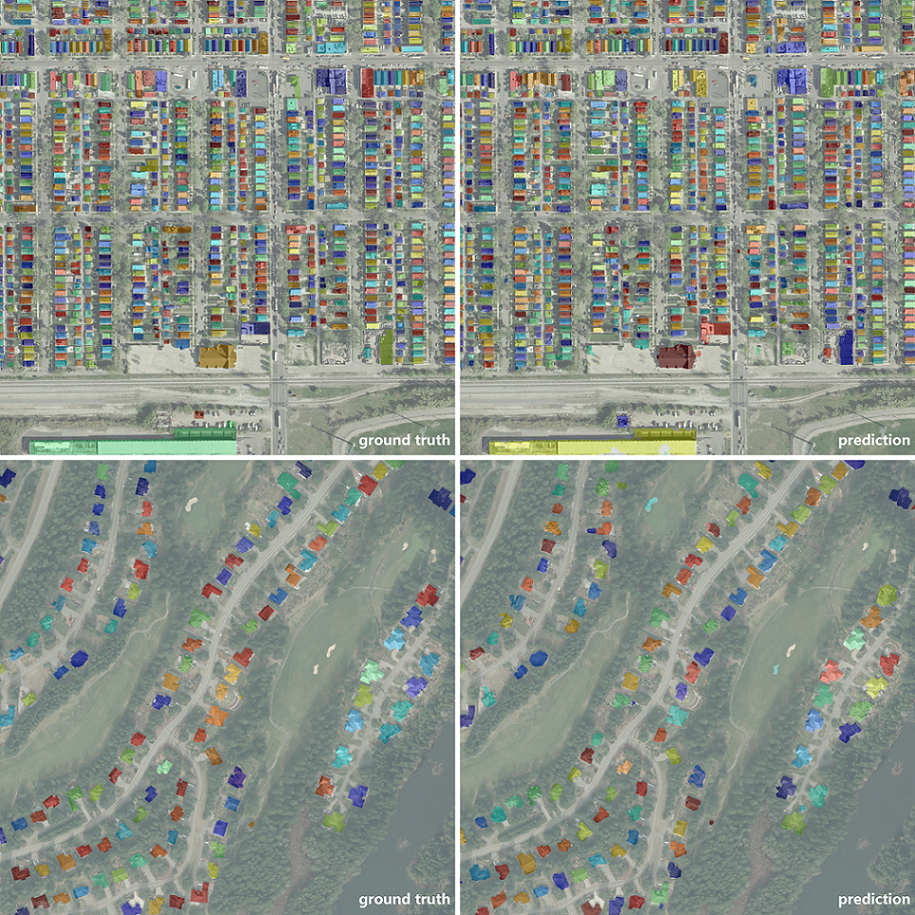}
\renewcommand{\figurename}{Fig}
\caption{\label{fig:inria_full} Segmentation results of two large-scale regions in Chicago (top) and Kitsap County (bottom). Colored areas mean building footprints, and different colors indicate different building instances. In this way, we can clearly see the performance of the network at instance level.}
\end{figure*}


\par
We compare the proposed RiFCN with the state-of-the-art in the literature~\cite{FCN,Maggiori17,SegNet,SegNet2,Bischke17,He17,Ohleyer18} on the Inria Aerial Image Labeling Data Set (cf. Table~\ref{tab:inria}). As on the ISPRS Potsdam data set, FCN~\cite{FCN} and SegNet~\cite{SegNet,SegNet2} are included. Moreover, in~\cite{Maggiori17}, FCN-MLP has been introduced, which upsamples and concatenates all intermediate feature maps of the convolutional component of a FCN and makes use of an MLP to reduce the concatenated features to predict segmentation maps. The authors of~\cite{Maggiori17} also provide the results of FCN-Skip, which creates multiple segmentation maps from different convolutional layers (at different resolutions), interpolates them to match the highest resolution, and adds the results to create the final semantic segmentation map. Bischke et al.~\cite{Bischke17} propose a cascaded multi-task learning SegNet (which we will call Multi-task SegNet hereafter) that addresses the problem of building segmentation by exploiting not only semantic segmentation masks but also geometric information (i.e., signed distance), aiming at preserving semantic boundaries in segmentation maps as far as possible. It is worth noting that compared to our RiFCN, Multi-task SegNet uses extra supervision information. He et al.~\cite{He17} propose a general framework, called Mask Region-based CNN (Mask R-CNN), which is capable of efficiently detecting objects in an image while simultaneously generating a segmentation mask for each detected instance. Mask R-CNN has proved its efficiency in computer vision. Later, the author of~\cite{Ohleyer18} makes use of Mask R-CNN for the building segmentation on satellite images.
\par
Table~\ref{tab:inria} shows results of the aforementioned methods on the Inria Aerial Image Labeling Data Set. It can be seen that the proposed RiFCN outperforms FCN, FCN-Skip, and FCN-MLP, and increments of overall IoU are 20.18\%, 11.03\%, and 9.33\%, respectively. This indicates that our feature fusion strategy (i.e. backward stream in our network) is more powerful and effective. In addition, compared with SegNet, the improvement in overall IoU achieved by RiFCN is 3.86\%. It is noteworthy that our network can even outperform Multi-task SegNet, which uses more supervision information for network learning. When comparing RiFCN against the recently proposed Mask R-CNN, we can observe an improvement of 14.47\% in IoU. Overall, the results show that the approach that produces high resolution segmentation map plays a crucial role for semantic segmentation tasks, and our method, which makes use of autoregressive recurrent connections in a bidirectional network architecture, can offer better results as compared to FCN-based and encoder-decoder methods. Fig.~\ref{fig:inria_full} shows segmentation results of two large-scale regions in Chicago and Kitsap County. Note that colored areas in Fig.~\ref{fig:inria_full} mean building footprints, and different colors indicate different building instances. In this way, we can clearly see the performance of the network at instance level.
\par
\begin{figure*}[!t]
\centering
\includegraphics[width=\linewidth]{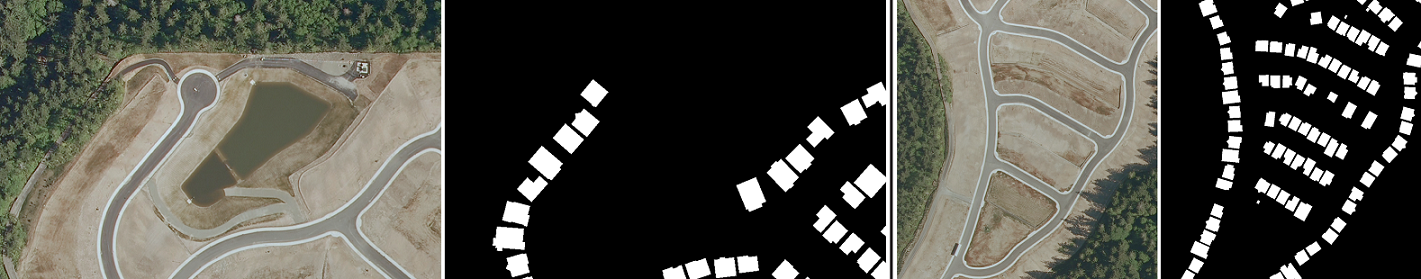}
\renewcommand{\figurename}{Fig}
\caption{\label{fig:bad_labels} Examples of ground truth labeling errors in the Inria Aerial Image Labeling Data Set.}
\end{figure*}

Furthermore, in our experiments, we noticed that there are some inaccuracies of the ground truth data, such as those shown in Fig.~\ref{fig:bad_labels}. Obviously, these inaccuracies affect the accurate evaluation of segmentation methods.

\section{Conclusion}
\label{sec:con}
In this paper, we propose a novel network architecture, RiFCN, for semantic segmentation of high resolution remote sensing data. In particular, the proposed network is composed of two parts, namely forward stream and backward stream. The forward stream is responsible for extracting multi-level convolutional feature maps from the input. And we design a reverse process (i.e., backward stream), which uses a series of autoregressive recurrent connections to hierarchically and progressively absorb high-level semantic features and render pixel-wise, high resolution predictions. By doing so, boundary-aware feature maps and high-level features are orderly embedded into the framework. Experiments demonstrate that the feature fusion strategy of the proposed RiFCN performs favorably against others (e.g., FCN, FCN-Skip, and FCN-MLP). In addition, compared to other network architectures such as SegNet and Mask R-CNN, the proposed network can offer better segmentation results for high resolution aerial imagery.

\section*{Acknowledgements}
The authors would like to thank the ISPRS for making the Potsdam data set available. In addition, they also would like to thank the INRIA Sophia-Antipolis Mediterranee for providing the Inria Aerial Image Labeling Data Set.
\par
This work is jointly supported by the China Scholarship Council, the European Research Council (ERC) under the European Union¡¯s Horizon 2020 research and innovation programme (grant agreement No [ERC-2016-StG-714087], Acronym: \textit{So2Sat}), and Helmholtz Association under the framework of the Young Investigators Group ``SiPEO'' (VH-NG-1018, \url{www.sipeo.bgu.tum.de}).

\section*{References}
\bibliographystyle{elsarticle-num}
\bibliography{rifcn}

\end{document}